\documentclass[preprint,12pt]{elsarticle}

\usepackage{amssymb}
\usepackage{amsmath}

\usepackage[T1]{fontenc}
\usepackage{utilsj, url}
\usepackage[ruled]{algorithm2e}
\usepackage{xcolor}
\usepackage{booktabs}

\usepackage{mathtools}
\usepackage{tikz}
\usepackage{pgfplots}
\pgfplotsset{width=13cm,compat=1.9, height=5cm}
\usetikzlibrary{decorations.markings, positioning,angles,quotes,arrows,shapes}

\newcommand{\E}{\mathbb{E}}
\DeclareMathOperator*{\argmax}{arg\,max}

\SetCommentSty{mycommfont}

\DeclareMathOperator*{\TN}{TruncatedNormal}

\DeclareMathOperator*{\TS}{TruncatedStudentT}

\DeclareMathOperator*{\ub}{ub}

\journal{Neural Networks}

\begin{document}

\begin{frontmatter}



\title{Deep Belief Markov Models for POMDP Inference}


\author[label1]{Giacomo Arcieri} 
\ead{giacomo.arcieri@ibk.baug.ethz.ch}
\author[label2]{Konstantinos G. Papakonstantinou}
\ead{kpapakon@psu.edu}
\author[label3]{Daniel Straub}
\ead{straub@tum.de}
\author[label1]{Eleni Chatzi}
\ead{chatzi@ibk.baug.ethz.ch}

\affiliation[label1]{organization={Institute of Structural Engineering, ETH Z{\"u}rich},
            city={Z{\"u}rich},
            postcode={8093}, 
            country={Switzerland}}

\affiliation[label2]{organization={Dept. of Civil and Environmental Engineering, The Pennsylvania State University},
            city={University Park},
            postcode={16802}, 
            state={PA},
            country={USA}}

\affiliation[label3]{organization={Engineering Risk Analysis Group \& Munich Data Science Institute, Technical University of Munich},
            city={Munich},
            postcode={80333}, 
            country={Germany}}

\begin{abstract}
This work introduces a novel deep learning-based architecture, termed the Deep Belief Markov Model (\emph{DBMM}), which provides efficient, model-formulation agnostic inference in Partially Observable Markov Decision Process (POMDP) problems. The POMDP framework allows for modeling and solving sequential decision-making problems under observation uncertainty. In complex, high-dimensional, partially observable environments, existing methods for inference based on exact computations (e.g., via Bayes' theorem) or sampling algorithms do not scale well. Furthermore, ground truth states may not be available for learning the exact transition dynamics. \emph{DBMMs} extend deep Markov models into the partially observable decision-making framework and allow efficient belief inference entirely based on available observation data via variational inference methods. By leveraging the potency of neural networks, \emph{DBMMs} can infer and simulate non-linear relationships in the system dynamics and naturally scale to problems with high dimensionality and discrete or continuous variables. In addition, neural network parameters can be dynamically updated efficiently based on data availability. \emph{DBMMs} can thus be used to infer a belief variable, thus enabling the derivation of POMDP solutions over the belief space. We evaluate the efficacy of the proposed methodology by evaluating the capability of model-formulation agnostic inference of  \emph{DBMMs} in benchmark problems that include discrete and continuous variables.\footnote{GitHub code available at \url{https://github.com/giarcieri/dmm-for-pomdps}.}
\end{abstract}


\end{frontmatter}


\section{Introduction}
\label{sec:intro}

Partially Observable Markov Decision Processes (POMDPs) \cite{sondik1971optimal,cassandra1998survey,spaan2005perseus} offer the potential to model and solve complex sequential decision-making problems under observation uncertainty. Within the POMDP framework, an agent receives an observation from the environment, which constitutes a partial and/or noisy signal of the environment's true state. The latter is hidden from the agent and, thus, is termed the \textit{hidden state}. Based on the received observation, the agent (implicitly or explicitly) forms a \textit{belief} over the hidden state and accordingly executes an action, which will affect the environment's next state. The agent receives a reward (or, equivalently, a cost) as a function of the action and the objective is to define a \textit{policy}, i.e., a sequence of actions, which maximizes/minimizes the total rewards/costs over a prescribed horizon.

The POMDP model includes the knowledge of the parameters, variables, and the structural form of the underlying transition dynamics and observation generating process, namely the models that describe the (stochastic) system's evolution of hidden states and the generation of observations from the system. Solution algorithms, e.g., planning algorithms \cite{ross2008online} or Reinforcement Learning (RL) methods \cite{arulkumaran2017deep}, generally assume availability of, or highly benefit from, explicit knowledge of the POMDP model \cite{matsuo2022deep}. However, the POMDP model is often not available and must be inferred from data \cite{papakonstantinou2014planningb,arcieri2023bridging}. 

We identify three main motivations for POMDP inference. The first motivation is for \textit{generative} purposes. The POMDP model inference may be necessary for building a simulator of the environment. This serves to implement, for instance, reinforcement learning solutions via interactions with the simulated environment \cite{rupprecht2022survey}, often essential in many engineering applications and/or safety-critical domains. The second motivation is to enable the inference of the \textit{belief} variable, namely a probability distribution over hidden states. Prior work \cite{arcieri2023pomdp} has shown that inferring a belief and applying RL algorithms on the belief space significantly improves the solution when compared against the alternative of directly applying RL algorithms on the observation and action history. The superiority of such a belief-based approach was demonstrated even under coupling of RL solutions with more complex neural network architectures, such as LSTMs or Transformers, that are suited for time-series input structures. Lastly, the inference of a model allows for full \textit{model-based RL} \cite{moerland2023model} solutions for POMDPs by coupling the learned model with planning algorithms that propagate the predicted beliefs over future time-steps.

While it can be beneficial to infer a model of the environment, the topic has received little attention in the literature and best practices are not generally available. Arcieri et al. \cite{arcieri2023bridging} propose a Hamiltonian Monte Carlo (HMC) sampling of a Hidden Markov Model (HMM) conditioned on actions to learn the parameters of the transition dynamics and observation generating process of the underlying POMDP problem entirely from collected data of observations and actions. However, this proposed methodology is not generally “model-formulation agnostic”, i.e., a structural
form for the transition dynamics and observation generating process is assumed, while the inference based on HMC sampling, similarly to all Markov Chain Monte Carlo (MCMC) methods, is computationally expensive. Within this framework, sequential/online model updates are also more difficult than approaches based on neural networks, which naturally allow such updates via stochastic gradient descent. Similarly, Lathourakis et al. \cite{lathourakis2023inference} apply HMC sampling to infer the parameters of the POMDP model, thus enabling the belief to be passed as input to a deep RL agent. However, this approach is also similarly characterized by the aforementioned issues.

A broad suite of deep learning-based approaches have been proposed for the model inference of the environment \cite{schmidhuber2015deep}. These are often related to model-based RL solutions, with a focus on fully observable
problems rather than on POMDPs \cite{nagabandi2018neural,chua2018deep,janner2019trust,wang2019benchmarking,arcieri2021model}. Relevant prior work on the POMDP problem includes the QMDP-net \cite{karkus2017qmdp}, which learns a small-scale, discrete representation of the original POMDP transition and observation model through recurrent networks, used then for planning via the QMDP method. The small-scale, discrete representation might pose a too strong and limiting assumption in complex, high-dimensional, and/or continuous problems, which is further limited by the analytical update of the beliefs. Igl et al. \cite{igl2018deep} delivered one of the first works to apply deep model-based reinforcement learning to POMDPs. The POMDP inference is based on variational learning, with the belief updated via use of a particle filter. Wang et al. \cite{wang2019robust} design a neural network-based scheme to infer the transition model of a POMDP problem and, in turn, approximate the belief state to deal with incomplete and noisy observations. The observation model in that work is not inferred, but assumed known a priori. None of these prior works nor any other work from the related literature applies structured inference methods that fully align with the POMDP structure. As explained next, Deep Markov Models (DMMs) \cite{krishnan2017structured} offer such an opportunity.

DMMs \cite{krishnan2017structured,girin2020dynamical} form a class of generative models, termed Gaussian state-space models, that preserve the Markovian structure of HMMs and employ neural networks to handle high-dimensional data and learn non-linear transition dynamics and observation generating processes, with parameters learned via variational inference. DMMs have been successfully implemented for applications across domains, including dynamical systems \cite{liu2022physics,bacsa2023symplectic,liu2023modelbased}, healthcare \cite{ozyurt2021attdmm}, audiovisual speech \cite{sadok2024multimodal}, or finance \cite{ferreira2020reinforced}.

DMMs do suffer certain limitations, as a result of their formulation following the HMM structure. They can account for actions, albeit only when these are fed as inputs to the system and not in the more structured format of a typical decision-support framework \cite{krishnan2017structured}. More importantly, they do not explicitly include belief inference, which is required for implementing a POMDP approach. In this work, we originally extend the DMMs for use within the POMDP framework by formally accounting for actions and by additionally shifting the focus to the belief inference. We term this novel model \textit{Deep Belief Markov Model} (\textit{DBMM}). \textit{DBMMs} leverage neural networks and variational inference for a model-formulation agnostic inference of the POMDP environment that scales even to continuous, non-linear, high-dimensional, and/or multi-component problems. \textit{DBMMs} can be used (i) as a generative simulator of the POMDP problem, (ii) for belief inference, which can be subsequently passed as input to classical model-free RL algorithms for enhancing performance, and/or (iii) for a fully model-based RL solution by propagating the learned belief for planning, while quantifying the estimation uncertainty via variational inference. Additionally, the model parameters can be dynamically updated based on data availability and can be efficiently adapted under system changes. 

We investigate the inference capability of the \emph{DBMM} on three benchmark problems. A first one comprises discrete variables for which the true beliefs can be exactly computed; a second one comprises continuous variables for which the exact beliefs are not available; and a third one constituting a real-world railway application problem derived from collected monitoring data, which comprises both discrete and continuous variables.  

The remainder of this paper is organized as follows. Section \ref{sec:preliminaries} overviews the necessary preliminaries on POMDPs, the concept of beliefs, and the fundamental description of DMMs for completeness. Section \ref{sec:bdmm} introduces the proposed Deep Belief Markov Model and presents a comparison to existing models, highlighting key differences. Section \ref{sec:experiments} investigates the inference capability of the \emph{DBMM} on the defined POMDP benchmarks. Finally, Section \ref{sec:conclusion} concludes the paper and outlines future work.

\section{Preliminaries}
\label{sec:preliminaries}
POMDPs \cite{kaelbling1998planning,papakonstantinou2014planning} form a generalization of Markov Decision Processes (MDPs), where states are hidden from the agent and can only be accessed through noisy and/or partially informative observations. A POMDP is defined by the tuple $\langle \set{S},\set{A},\set{O},R,\mathbb{P},\mathbb{O},b_0,\Omega_b,T,\gamma\rangle$ where $\set{S}$, $\set{A}$, and $\set{O}$ are the hidden state, action, and observation spaces, $R:\set{S} \times \set{A}  \rightarrow\mathbb{R}$ is the reward function of the problem that outputs the reward $r_t=R(s_t, a_t)$ at time-step $t$, $\mathbb{P}:\set{S}\times\set{S}\times\set{A}\rightarrow p(s_{t+1}|s_t,a_t)$ is the process defining the transition dynamics, $\mathbb{O}:\set{S} \times \set{A} \times \set{O}  \rightarrow p(o_t|s_t,a_{t-1})$ is the observation generating process, $b_0\in\Omega_b$ is the initial belief (i.e., the belief on the initial hidden state $s_0\in\set{S}$), $T$ is the final time-step, i.e., the horizon of the problem, and $\gamma$ is the discount factor. 

The objective of the POMDP is to determine the optimal policy $\pi^*$ that maximizes the expected sum of rewards:
\begin{equation}\label{eq:obj_pomdp}
    \pi^*=\argmax_\pi\E\left[\sum_{t=0}^T\gamma^tr_t\right]
\end{equation}
As the states are not observable, the agent is required to find a (non-Markovian) policy based on the entire history of observations and actions, i.e., $a_t=\pi(o_0, a_0,\dots, o_t)$. As such, the policy space grows exponentially with the horizon $T$, known as the \textit{curse of history} \cite{silver2010monte}. To alleviate this curse, the belief variable has been introduced in the POMDP framework. This defines a probability distribution over the hidden states given the entire history of past observations and actions, $b_t=p(s_t \mid o_{0:t}, a_{0:t-1})$. The belief encodes the agent's knowledge in a compact representation that summarizes the past history to estimate the probable hidden states. It is a sufficient statistic of the history of observations and actions \cite{pineau2006anytime}. Simply put, knowledge of $b_t$ provides the decision-maker with the same amount of information as the complete history, albeit with reduced dimensionality. As a result, the original POMDP is transformed into the so-called equivalent \emph{belief-MDP}. Under this premise, the agent can plan the next action by exclusively relying on the current belief, i.e., the planning of the action $a_t=\pi(b_t)$ is Markovian. The belief $b$ is defined over a domain $\Omega_b$, which depends on the hidden state space $\set{S}$. For example, if the hidden states are discrete, the belief is defined over a continuous $|\set{S}|-1$ dimensional simplex, namely $\Omega_b=[0,1]^{|\set{S}|}$ where one dimension is redundant because all dimensions sum up to 1. If the hidden states are continuous, one can generally assume $\Omega_b=\mathbb{R}^{|\set{S}|}$.

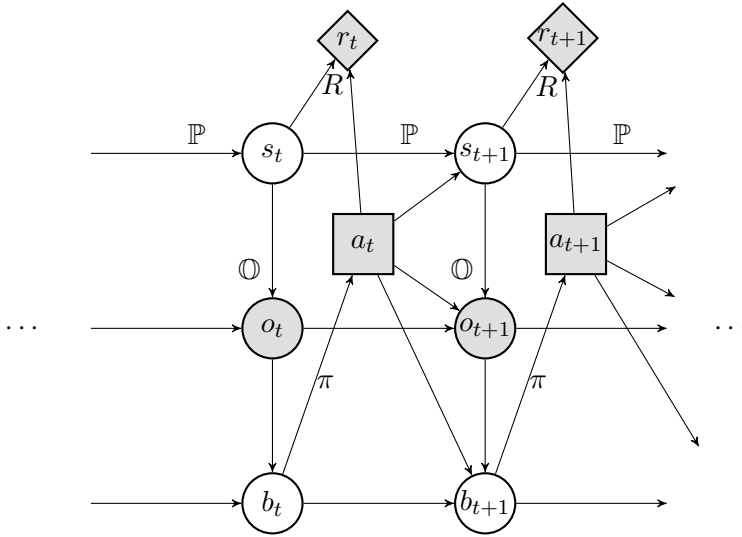
\begin{figure}[htb]
\begin{tikzpicture}[auto,node distance=8mm,>=latex,font=\small]
\tikzset{
    >=stealth',
    node distance=1.5cm,
    state/.style={minimum size=50pt,font=\small,circle,draw},
    dots/.style={state,draw=none},
    edge/.style={->},
    trans/.style={font=\footnotesize,above=2mm},
    reflexive/.style={out=120,in=60,looseness=5,relative},
    squared/.style={rectangle, draw=black, fill=gray!25, thick, minimum size=8mm},
    round/.style={circle, draw=black, fill=white, thick, minimum size=8mm},
    round_hidden/.style={circle, draw=black, fill=gray!25, thick, minimum size=8mm},
    decision/.style={diamond, draw=black, fill=gray!25, thick, minimum size=8mm},
    minimum size=8mm,inner sep=0pt
  }
    
    \node[round] (b0) {$b_t$};
    \node[round_hidden] (o0) [above=15mm of b0] {$o_t$};
    \node[round] (s0) [above=15mm of o0] {$s_t$};
    \node [dots]  (d0)  [left=20mm of b0] {};
    \node [dots]  (ds0)  [left=20mm of s0] {};
    \node [dots]  (do0)  [left=20mm of o0] {$\cdots$};
    \node[squared] (a0) [below right=5mm and 5mm of s0] {$a_t$};
    \node[decision] (r0) [above right=10mm and 5mm of s0] {$r_t$};
    \node[round,right=20mm of b0] (b1) {$b_{t+1}$};
    \node[round_hidden] (o1) [above=15mm of b1] {$o_{t+1}$};
    \node[round] (s1) [above=15mm of o1] {$s_{t+1}$};
    \node[squared] (a1) [below right=5mm and 5mm of s1] {$a_{t+1}$};
    \node[decision] (r1) [above right=10mm and 5mm of s1] {$r_{t+1}$};
    \node [dots]  (d1)  [right=20mm of b1] {};
    \node [dots]  (ds1)  [right=20mm of s1] {};
    \node [dots]  (do1)  [right=20mm of o1] {$\cdots$};
    
    \draw[->] (d0)--(b0);
    \draw[->] (ds0)to node[above right=-1.5mm and 0.mm]{ $\mathbb{P}$}(s0);
    \draw[->] (b0)--(b1);
    \draw[->] (s0)to node[above right=-1.5mm and 0.mm]{ $\mathbb{P}$}(s1);
    \draw[->] (s0)to node[below left=0mm and -1mm]{ $\mathbb{O}$}(o0);
    \draw[->] (do0)--(o0);
    \draw[->] (o0)--(b0);
    \draw[->] (o0)--(o1);
    \draw[->] (b0)to node[above right=-5mm and -3mm]{ $\pi$}(a0);
     \draw[->] (s0)--(r0);
    \draw[->] (a0)--(b1);
    \draw[->] (a0)--(s1);
    \draw[->] (a0)--(o1);
     \draw[->] (a0)to node[above left=3.5mm and -1mm]{ $R$}(r0);
    \draw[->] (b1)--(d1);
    \draw[->] (s1)to node[above right=-1.5mm and 0.mm]{ $\mathbb{P}$}(ds1);
    \draw[->] (s1)to node[below left=0mm and -1mm]{ $\mathbb{O}$}(o1);
    \draw[->] (o1)--(b1);
    \draw[->] (o1)--(do1);
    \draw[->] (b1)to node[above right=-5mm and -3mm]{ $\pi$}(a1);
    \draw[->] (s1)--(r1);
    \draw[->] (a1)to node[above left=3.5mm and -1mm]{ $R$}(r1);
    \draw[->] (a1)--(d1);
    \draw[->] (a1)--(ds1);
    \draw[->] (a1)--(do1);
\end{tikzpicture}
\caption{Probabilistic graphical model of a POMDP.}
\label{fig:pomdp}
\end{figure}

A POMDP can be represented as a probabilistic graphical model \cite{koller2009probabilistic}, as shown in Figure \ref{fig:pomdp}, where shaded nodes denote observed variables, while edges encode the dependence structure among variables.

\subsection{Belief update}
\label{sec:belief_update}

When a new observation is available, the belief is updated via Bayes' theorem, namely for discrete hidden states the posterior belief is computed as follows:
\begin{equation}\label{eq:belief_discrete}
b_{t+1}=\frac{p(o_{t+1}\mid s_{t+1},a_t)\sum_{s_t\in \set{S}}p(s_{t+1}\mid s_t,a_t)b_t}{\eta}
\end{equation}
and for continuous hidden states as:
\begin{equation}\label{eq:belief_continuous}
b_{t+1}=\frac{p(o_{t+1}\mid s_{t+1},a_t)\int p(s_{t+1}\mid s_t,a_t)b_t \,ds_t}{\eta}
\end{equation}
where $\eta$ is a normalizing factor, i.e., a sum/integral of the numerator over $s_{t+1}\in \set{S}$. The right term of the expression propagates the posterior belief, $b_t$, to the next time-step by means of the transition model, $\mathbb{P}=p(s_{t+1}\mid s_t,a_t)$, resulting in the prior belief at the next time-step $\tilde{b}_{t+1}$. We introduce the \textit{belief transition} operator $\mathcal{T}: \Omega_b\times\set{A}\rightarrow\Omega_b$, which is a function that propagates the posterior belief, $b_t$, to the prior belief at the next time-step, $\tilde{b}_{t+1}$, based on the chosen action and the transition dynamics, namely:
\begin{equation}
    \tilde{b}_{t+1}=\mathcal{T}(b_t, a_t)=\sum_{s_t\in \set{S}}p(s_{t+1}\mid s_t,a_t)b_t
\end{equation}
for discrete hidden states, while an integral replaces the sum operator for continuous hidden states. We can, thus, rewrite the belief update of Equations \ref{eq:belief_discrete} and \ref{eq:belief_continuous} as:
\begin{equation}
b_{t+1}=\frac{p(o_{t+1}\mid s_{t+1},a_t)\tilde{b}_{t+1}}{\eta}
\end{equation}

We can correspondingly introduce the \textit{belief inference} operator $\mathcal{Q}:\Omega_b\times\set{O}\rightarrow\Omega_b$, which employs the newly received observation and the observation model in order to update the prior belief, $\tilde{b}_{t+1}$, to the posterior belief, $b_{t+1}$ in that time-step:
\begin{equation}
b_{t+1}= \mathcal{Q}(\tilde{b}_{t+1}, o_{t+1})=\mathcal{Q}(\mathcal{T}(b_t, a_t), o_{t+1})
\end{equation}

\subsection{Deep Markov Models}
\label{sec:dmm}
DMMs are composed of two main models, the \textit{generative} model, with a set of parameters denoted by $\theta$, and the \textit{inference} model, with a set of parameters denoted by $\phi$. The generative model is, in turn, composed of two sub-models, approximated via use of neural network functions, whose parameters we denote as $\theta=(\omega, \kappa)$. These parameterize the transition model $p_\omega(s_t\mid s_{\omega_{t-1}})$ and the observation model $p_\kappa(o_t\mid s_{\omega_t})$, with $s_{\omega_t}\sim p_\omega(s_t\mid s_{\omega_{t-1}})$, where the subscript is here used to indicate that the variable is estimated by the model. Exploiting the Markov property, one can factorize the joint probability of observations and hidden states in time as:
\begin{equation}\label{eq:generative_dmm}
    p_\theta(o_{1:T}, s_{1:T})=\prod_{t=1}^Tp_\kappa(o_t\mid s_{\omega_t})p_\omega(s_t\mid s_{\omega_{t-1}})
\end{equation}
Equation \ref{eq:generative_dmm} assumes that the initial state $s_0$ is known. When this is not the case, a distribution $p(s_0)$ can be assumed and multiplied to the right-hand side.

The posterior probability $p_\theta(s_{1:T}\mid o_{1:T})$ is typically intractable and, within the framework of variational inference, can be approximated by the inference model that learns the approximate posterior distribution $q_\phi(s_{1:T}\mid o_{1:T})$. Exploiting the fact that $s_t\perp \!\!\! \perp o_{1:t-1} \mid s_{t-1}$ from the hidden Markov model structure, one can write:
\begin{equation}\label{eq:inference_dmm}
    q_\phi(s_{1:T}\mid o_{1:T})=\prod_{t=1}^Tq_\phi(s_t\mid s_{\phi_{t-1}}, o_{t:T})
\end{equation}
As evident from Equation \ref{eq:inference_dmm}, the original DMM from Krishnan et al. \cite{krishnan2017structured} accesses the future observations $o_{t:T}$ to infer the current hidden state $s_t$. This is admissible in a predictive context of smoothing filters, which access information from the latter samples of a batch in order to improve the estimates within the batch. However, $o_{t:T}$ is not available in a RL/POMDP context, where sequential decisions are made only based on past information. Krishnan et al. \cite{krishnan2017structured} employ two neural networks for the inference model $(\psi,\zeta)$. Namely, a Recurrent Neural Network (RNN) to extract the recurrent hidden states $h_t$ (not to be confused with the HMM hidden states $s_t$), and a Combiner network, which combines $h_t$ and $s_{t-1}$ to infer $s_t$. The DMM structure is summarized in Figure \ref{fig:dmm}.

\begin{figure}[t]
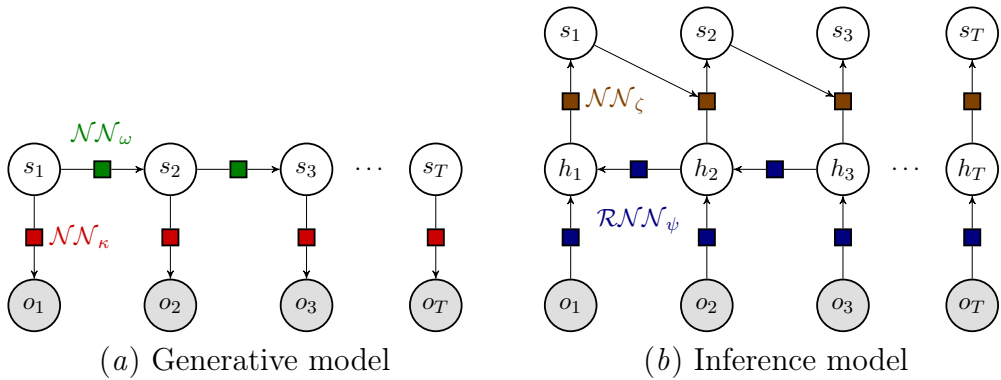

\floatconts
  {fig:dmm}
  {\caption{Graphical representation of the Deep Markov Model. The generative model (left) is composed of two neural networks that learn the transition and the observation models. The inference model (right) is composed of a RNN that learns recurrent hidden states from (future) observations and a second neural network (termed Combiner) that infers the systems hidden states.}}
  {%
    \subfigure[Generative model]{\label{fig:dmm_gen}%
      \includeteximage[width=0.45\textwidth]{dmm_generative}
      }%
    \qquad
    \subfigure[Inference model]{\label{fig:dmm_inf}%
      \includeteximage[width=0.45\textwidth]{dmm_inference}}
  }
\end{figure}

The parameters of the DMM are ideally learned by maximizing the log-marginal likelihood $p_\theta(o_{1:T})$. This can be decomposed into an approximation of the log-marginal likelihood through parameters $\phi$ and the Kullback-Leibler (KL)
divergence between the approximate posterior distribution $q_\phi(s_{1:T}\mid o_{1:T})$ and the true posterior distribution $p_\theta(s_{1:T}\mid o_{1:T})$, namely:
\begin{equation}
    \log p_\theta(o_{1:T})=\mathcal{L}(o_{1:T}; (\theta, \phi))+ D_{\mathrm{KL}}\left(q_\phi(s_{1:T}\mid o_{1:T})\parallel p_\theta(s_{1:T}\mid o_{1:T})\right)
\end{equation}
where the right term is intractable because of the non-linear relationship between $s_{1:T}$ and $o_{1:T}$. Nevertheless, by noticing that 
\begin{equation}
    D_{\mathrm{KL}}\left(q_\phi(s_{1:T}\mid o_{1:T})\parallel p_\theta(s_{1:T}\mid o_{1:T})\right)\geq 0
\end{equation}
one can maximize $\mathcal{L}(o_{1:T}; (\theta, \phi))$; this is termed the Variational Lower Bound (VLB) or Evidence Lower Bound (ELBO). Following the previous factorization, this can be defined as:
\begin{equation}\label{eq:vlb_dmm}
    \mathcal{L}(o_{1:T}; (\theta, \phi)) = \sum_{t=1}^T\E_{q_\phi}\left[\log p_\kappa(o_t\mid s_{\omega_t}) - D_{\mathrm{KL}}\left(q_\phi(s_t\mid s_{\phi_{t-1}}, o_{t:T})\parallel p_\omega(s_t\mid s_{\omega_{t-1}})\right) \right]
\end{equation}
Krishnan et al. \cite{krishnan2017structured} compute the right term of Equation \ref{eq:vlb_dmm} analytically following the Gaussian likelihood assumption, while the expectation is computed via Monte Carlo estimation.

\section{Deep Belief Markov Models}
\label{sec:bdmm}

The DBMM extends the DMM by more explicitly incorporating actions and including beliefs in order to assimilate the structure of the POMDP graphical model (Figure \ref{fig:pomdp}). In particular, the belief inference of the \emph{DBMM} mirrors the process of belief inference via a Bayesian update, as described in Equations \ref{eq:belief_discrete} and \ref{eq:belief_continuous}, where operations are now approximated with neural networks, as explained subsequently.

Similar to the DMM, the \emph{DBMM} is also composed of a generative $\theta$ and an inference model $\phi$. The generative model is, in turn, composed of two neural networks $\theta=(\kappa, \omega)$. Different to the DMM, the neural network $\omega$ does not approximate the transition model, namely the evolution of hidden states conditioned on the actions, but the operator $\mathcal{T}$ introduced in Section \ref{sec:belief_update} to describe the evolution of the belief state, namely:
\begin{equation}
    \tilde{b}_{\omega_t}=\mathcal{T}_\omega(\tilde{b}_{\omega_{t-1}}, a_{t-1})=p_\omega(s_t\mid \tilde{b}_{\omega_{t-1}}, a_{t-1})
\end{equation}
Hence, $\omega$ is a \textit{belief transition} model, which learns to propagate the belief in absence of information from observations, namely it learns the prior $\tilde{b}_t$. In the generative model, the hidden state $s_{\omega_t}$ is sampled from the prior belief distribution $\tilde{b}_{\omega_{t}}$, i.e., $s_{\omega_t}\sim\tilde{b}_{\omega_{t}}$, where the subscript, $_{\omega_t}$, denotes that this is an \textit{estimated} hidden state, namely learned by the model, and not the ground truth (actual) hidden state $s_t$. 

The second neural network of the generative model $\kappa$ parameterizes the observation model $p_\kappa(o_t\mid s_{\omega_t}, a_{t-1})$, as in the DMM (the observation may also depend only on the hidden state and not on the action). Thus, the factorized joint probability of the generative model can be computed as:
\begin{equation}\label{eq:generative_bdmm}
    p_\theta(o_{1:T}, s_{1:T}\mid a_{1:T})=\prod_{t=1}^Tp_\kappa(o_t\mid s_t, a_{t-1})p_\omega(s_t\mid \tilde{b}_{\omega_{t-1}}, a_{t-1})
\end{equation}
where $\tilde{b}_{\omega_0}$ is the prior on the initial condition $p(s_0)$.

The inference model $\phi$ is composed of two neural networks $\phi=(\omega, \psi)$. The first network parameterizes the operator $\mathcal{T}$ and shares the parameters $\omega$ with the neural network of the generative model. This propagates the posterior belief at time $t-1$ to the next time-step $t$, in absence of information from the observations, to compute the next prior belief:
\begin{equation}\label{eq:t_bdmm}
    \tilde{b}_{\phi_t}=\mathcal{T}_\omega(b_{\phi_{t-1}}, a_{t-1})=q_\omega(s_t\mid b_{\phi_{t-1}}, a_{t-1})
\end{equation}
The second neural network $\psi$ parameterizes the \textit{belief inference} operator $\mathcal{Q}$. It admits as input the current prior belief computed via $\mathcal{T}_\omega$ and estimates the posterior belief given the current observation $o_t$:
\begin{equation}\label{eq:inf_bdmm}
    b_{\phi_t}=\mathcal{Q}_\psi(\tilde{b}_{\phi_{t}}, o_{t})=q_\psi(s_t\mid \tilde{b}_{\phi_{t}}, o_{t})
\end{equation}
It should be noted that, in the \emph{DBMM} case, no future information (i.e., future actions and/or observations) is used for inference, in order to respect the sequential nature of the decision-making problem. As a result of the re-established Markov property, there is also no need to employ RNNs. The complete graphical model of the \emph{DBMM} is reported in Figure \ref{fig:bdmm}.

\begin{figure}[t]
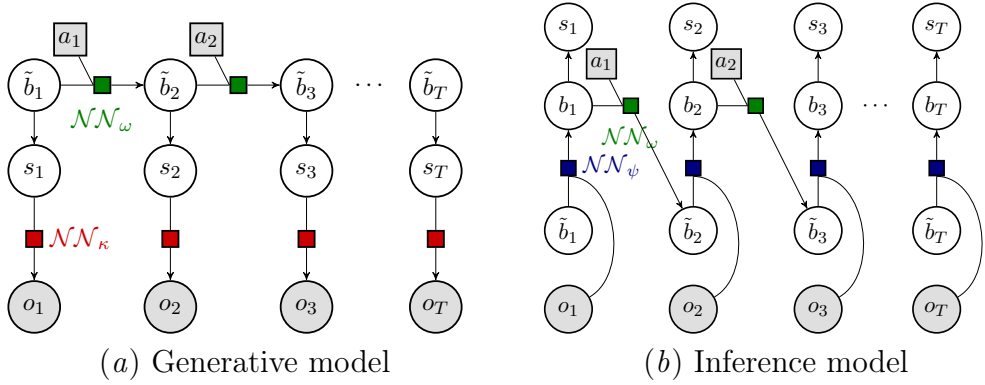

\floatconts
  {fig:bdmm}
  {\caption{Graphical representation of the Deep Belief Markov Model (\emph{DBMM}). The generative model (left) is composed of two neural networks that learn the belief transition operator and the observation model. The inference model (right) is composed of two neural networks that learn the belief transition and the belief inference operators.}}
  {%
    \subfigure[Generative model]{\label{fig:bdmm_gen}%
      \includeteximage[width=0.45\textwidth]{bdmm_generative}
      }%
    \qquad
    \subfigure[Inference model]{\label{fig:bdmm_inf}%
      \includeteximage[width=0.45\textwidth]{bdmm_inference}}
  }
\end{figure}

In the inference model, the \textit{network} hidden states are sampled from the estimated network posterior beliefs, i.e., $s_{\phi_t}\sim b_{\phi_t}$. The probability distribution of the inference model can thus be factorized as:
\begin{equation}\label{eq:inference_bdmm}
    q_\phi(s_{1:T}\mid o_{1:T}, a_{1:T})=\prod_{t=1}^Tq_\phi(s_t\mid b_{\phi_{t-1}}, a_{t-1}, o_{t})=\prod_{t=1}^Tq_\omega(s_t\mid b_{\phi_{t-1}}, a_{t-1})q_\psi(s_t\mid \tilde{b}_{\phi_{t}}, o_{t})
\end{equation}
Finally, the VLB of the \emph{DBMM} is:
\begin{equation}\label{eq:vlb_bdmm}
\begin{split}
    \mathcal{L}(o_{1:T}, a_{1:T}; (\theta, \phi)) =&\sum_{t=1}^T\E_{q_\phi}\Bigl[\log p_\kappa(o_t\mid s_{\phi_t}, a_{t-1}) - \\
    &D_{\mathrm{KL}}\left(q_\phi(s_t\mid b_{\phi_{t-1}}, a_{t-1}, o_{t})\parallel p_\omega(s_t\mid \tilde{b}_{\omega_{t-1}}, a_{t-1})\right) \Bigr]
\end{split}
\end{equation}
The left term maximizes the likelihood of the observations, while the right term ensures that the \emph{belief transition} model and the \emph{belief inference} model remain close. Thus, the latter acts as a regularizer, which combines generative and inference models into a joint training, enabling the parameters $\omega$ to learn a meaningful transition dynamics as informed by the posterior $q_\phi$. This objective function is computed via use of Monte Carlo method in this work. 

\subsection{Comparison with previous work and models}
\label{sec:comparison}

The \emph{DBMM} belongs to a class of models that can be defined as deep State Space Models (SSMs), also called Dynamical Variational Autoencoders (DVAEs) as codified by Girin et al. \cite{girin2020dynamical}, which combine neural networks with the SSM structure. When examining non-deep SSM variants, the Ensemble Kalman Filter (EnKF) \cite{Evensen2009,katzfuss2016understanding} offers a suitable option for application in POMDP problems that present continuous variables with possible nonlinear relationships. The EnKF stores a set of points (or particles), which approximate a distribution over the hidden state (i.e., the belief), which is updated when new observations become available. A substantial benefit of using the \emph{DBMM} over the EnKF is that the latter assumes prior knowledge of the ground truth model in order to update the belief, while in the \emph{DBMM} the POMDP model is learned. This benefit is further emphasized in real-world applications, where the ground truth model is often unknown, hindering the applicability of the EnKF but not of the \emph{DBMM}. Additionally, while the EnKF is generally limited to the gaussian assumption, the \emph{DBMM} is not subject to this constraint.

A comparison of the \emph{DBMM} against the DMM reveals a number of differences. First, the original DMM uses information from the future, i.e., $o_{t:T}$, to infer the hidden state $s_t$, which is why it is also referred to as Deep Kalman \emph{Smoother} in Krishnan et al. \cite{krishnan2017structured}. This generates important differences in the inference model architecture, such as the use of an RNN and a Combiner network. Using future information is infeasible in the RL/POMDP context. Thus, no future information is used and the \emph{DBMM} fully respects the Markov property of the belief-MDP, as visible when comparing the graphical models of Figure \ref{fig:pomdp} and \ref{fig:bdmm}. Furthermore, the DMM does not present an explicit belief inference. As a result, the DMM estimates hidden states $s_t$ given previous (estimated) hidden states. By contrast, the \emph{DBMM} estimates beliefs (and, hence, hidden states too) given previous (estimated) beliefs. This can be noted when comparing Equation \ref{eq:generative_bdmm} with Equation \ref{eq:generative_dmm} and Equation \ref{eq:inference_bdmm} with Equation \ref{eq:inference_dmm}. These differences lead in turn to a different VLB to train the models, see Equation \ref{eq:vlb_bdmm} and Equation \ref{eq:vlb_dmm}. As a result and differently from the DMM, the belief inference of the \emph{DBMM} fully reflects the belief updates via Bayes theorem, see Equations \ref{eq:belief_discrete}-\ref{eq:belief_continuous}, and finally forms a neural network approximation of the belief-MDP framework. Additionally, the (estimated) belief carries more information than the (estimated) hidden state, from which it is sampled at inference time, and as such it can allow better future belief predictions. 

From the family of deep SSMs, Bayer et al. \cite{bayer2014learning} propose the Stochastic Recurrent Network (STORN), which originally integrates a RNN into the SSM structure. However, this model presents some inconsistencies, as noted in Girin et al. \cite{girin2020dynamical}, which makes it a less interesting comparison. Hafner et al. \cite{hafner2019learning} propose the Recurrent State Space Model (RSSM), which is identical to the Variational Recurrent Neural Network (VRNN) originally proposed in Chung et al. \cite{chung2015recurrent}. This model learns a deterministic recurrent hidden state $h_t$ and a stochastic hidden state $s_t$ as part of the transition model. There are then several intricacies regarding $s_t$ and $h_t$ (e.g., in both the transition model, where $h_t$ depends on both $h_{t-1}$ and $s_{t-1}$, and the observation model), which do not respect the classical POMDP dependencies and make RSSM/VRNN very different from \emph{DBMM}. Another major difference is that the recurrent hidden state $h_t$ (as also the stochastic $s_t$) does not form a probability distribution with a specific physical meaning in the actual POMDP problem, but it is instead a network vector representation from the RNN. To the best of our knowledge, the \emph{DBMM} is the only model that fully reflects the POMDP/belief-MDP structure and the predicted beliefs have a clear, interpretable meaning, as they approximate the true POMDP beliefs and, therefore, form a probability distribution over the true hidden state space.

Outside of the family of deep SSMs, further works have proposed neural network-based models for the POMDP inference task \cite{karkus2017qmdp,igl2018deep,wang2019robust}. These models, however, present significant differences with the \emph{DBMM}, as already mentioned in Section \ref{sec:intro}.

\section{Experiments}
\label{sec:experiments}

This section explores the inference performance of the \emph{DBMM} over the POMDP benchmarks. In this work, we are only interested in evaluating the POMDP/belief inference capability of the model, and not in optimizing a policy; therefore, any assigned policy may be employed to generate the data. A random policy is assumed in all experiments in this paper.

\subsection{Discrete control benchmark}
\label{sec:disc_benchmark}

We present this discrete control benchmark for corroboration of the proposed approach, since in this case we can analytically compute the true posterior beliefs via Equation \ref{eq:belief_discrete}, which can be compared to the posterior beliefs estimated by the \emph{DBMM}. In a discrete control benchmark, where all involved variables, namely hidden states, actions, and observations, are discrete, solutions based on MCMC/HMC inference can work well \cite{arcieri2023bridging}. It should though be noted that the \emph{DBMM} can offer advantages in terms of dimensionality also in some discrete cases, e.g., systems with dependencies. 

The benchmark is taken from Papakonstantinou et al. \cite{papakonstantinou2018pomdp} and corresponds to a bridge maintenance planning setting with $|\set{S}|=5$, $|\set{A}|=4$, and $|\set{O}|=3$. Transition and observation models are reported in \ref{apd:disc}. All neural networks that compose the \emph{DBMM} (i.e., ($\omega, \kappa, \psi$)) learn categorical distributions and are composed of a single hidden layer of 100 neurons. A minimal hyper-parameter optimization was performed; a more thorough optimization would likely improve results further; however, this lies outside the scope of this work.

In our evaluation, we simulate the sequential learning and model updating steps that would be typical in a sequential decision-making problem, similarly to a deep model-based RL loop (except that the policy is not learned here). 500 trials, each of 100 time-steps (i.e., $T=100$), are sampled from the environment with a given policy (here random). For each trial, the environment is re-initialized with a different random seed. The \emph{DBMM} estimates the network beliefs $b_{\phi_t}$ on the basis of the observation $o_t$, the action $a_{t-1}$, and the previously estimated belief $b_{\phi_{t-1}}$. 
The \emph{DBMM} is then updated on the observations and actions collected across 500 trials by minimizing Equation \ref{eq:vlb_bdmm}. Afterwards, a new evaluation loop is initiated. The evaluation procedure is summarized in Algorithm \ref{alg:evaluation} in \ref{apd:alg}.

In assessing the accuracy of the predicted beliefs, we are interested in knowing whether these are as good in representing the hidden states as the true beliefs. Namely, we are interested in comparing the distance (respectively the closeness in the approximation) between the distribution of the predicted beliefs and the true hidden states with the distance between the distribution of true beliefs and the true hidden states. This is measured by the Cross-Entropy (CE) loss.

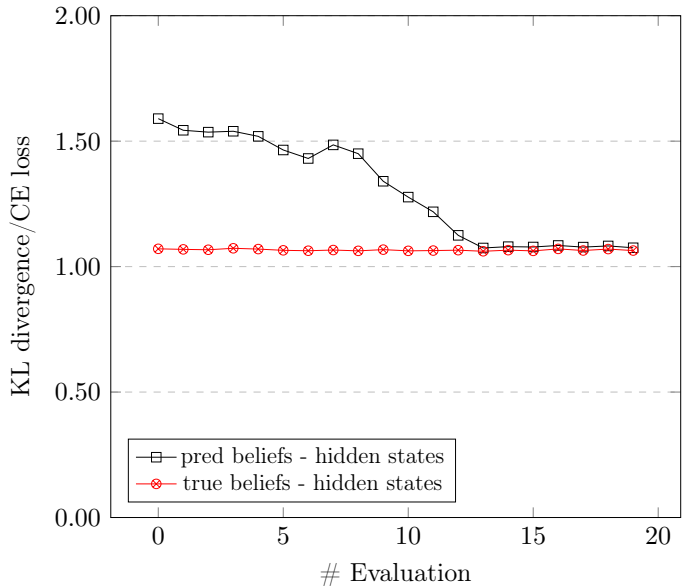
\begin{figure*}[ht]
    \centering
    \begin{tikzpicture}[scale=0.9]
    \begin{axis}[
        width=10cm,height=9cm,
        xlabel={\# Evaluation},
        ylabel={KL divergence/CE loss},
        label style={font=\small},
        ymin=-0., ymax=2.,
        ytick={0,0.5, 1.,1.5,2},
        tick label style={font=\small},
        legend pos=south west,
        ymajorgrids=true,
        grid style=dashed,
        legend style={nodes={scale=0.8, transform shape}}, 
        y tick label style={
            /pgf/number format/.cd,
            fixed,
            fixed zerofill,
            precision=2,
            /tikz/.cd
        }
    ]
    
    \addplot[
        color=black,
        mark=square,
        ]
        coordinates {
        (0,1.5897490978240967)
        (1,1.5433390140533447)
        (2,1.5358678102493286)
        (3,1.5394281148910522)
        (4,1.5191305875778198)
        (5,1.4647393226623535)
        (6,1.4309039115905762)
        (7,1.4851065874099731)
        (8,1.4500471353530884)
        (9,1.340147852897644)
        (10,1.2769622802734375)
        (11,1.2183810472488403)
        (12,1.124570369720459)
        (13,1.0742404460906982)
        (14,1.079526662826538)
        (15,1.0781564712524414)
        (16,1.084229588508606)
        (17,1.0777467489242554)
        (18,1.0824062824249268)
        (19,1.0750880241394043)
        };
    \addplot[
    color=red,
    mark=otimes,
    ]
    coordinates {
    (0,1.0708679545127506)
    (1,1.0686750144509427)
    (2,1.0671482827496768)
    (3,1.0729734545460745)
    (4,1.0698125799065124)
    (5,1.0648716000416956)
    (6,1.0632697375777906)
    (7,1.0660303636067765)
    (8,1.062803255034327)
    (9,1.0676326854259117)
    (10,1.062817808836702)
    (11,1.0639266076609726)
    (12,1.065465690685597)
    (13,1.0611091236644468)
    (14,1.0655575277939584)
    (15,1.0626665187190432)
    (16,1.070563076442391)
    (17,1.064158222842624)
    (18,1.0698801120438333)
    (19,1.0639298618739714)
    };
    \legend{pred beliefs - hidden states, true beliefs - hidden states}
        
    \end{axis}
    \end{tikzpicture}
    
    \caption{CE loss between the true beliefs and the hidden states (red) and between the predicted beliefs and the hidden states (black) over the evaluation loops in the discrete case (at each evaluation loop the two values of CE loss are computed over the same 500 trials, afterwards the model is updated on these trials).}
    \label{fig:discrete}
\end{figure*}

Figure \ref{fig:discrete} reports the CE loss between the true beliefs and the hidden states (red) and the CE loss between the predicted beliefs and the hidden states (black) over the evaluation loops. The \emph{DBMM} beliefs are randomly predicted by the model at the first evaluation run (before any model update). Their accuracy then consistently improves and their measured CE loss eventually converges to the CE loss of the true beliefs in a few model updates. As the (true) beliefs represent the best known representation of the hidden states in the POMDP framework (given the history of past observations and actions), this result implies that the beliefs predicted by the \emph{DBMM} are ultimately (almost) as good as the optimal representation, yet without assuming prior knowledge of the POMDP model.

As aforementioned in Section \ref{sec:comparison}, an appealing feature of the \emph{DBMM} is the interpretability of the results and, particularly, of the learned belief representation. Based on the predictions over 500 trials (unseen data) by the converged model, we inspect the accuracy per class of the estimated beliefs, i.e., the Multi-Class Accuracy (MCA), which provides more comprehensive information in cases where the different hidden state classes may not be equally represented in the evaluation data. Table \ref{tab:mca-discrete} reports the MCA, i.e., per-class accuracy score, of the predicted belief with respect to the true hidden states, compared with the MCA of the true beliefs (gold standard) again with respect to the true hidden states. Interestingly, the penultimate state is (almost) never correctly recovered neither by the true beliefs nor by the \emph{DBMM} estimated beliefs. The other states may be better recovered by the true beliefs or by the model, but with an overall similar amount of information carried by the two representations, as shown by the similar CE loss.

\begin{table}[hbtp]
\floatconts
  {tab:mca-discrete}
  {\caption{Multi-Class Accuracy (MCA), i.e., per-class accuracy score, of the true beliefs with respect to the true hidden states (top row) and MCA of the predicted belief with respect to the true hidden states (bottom row).}}
  {\begin{tabular}{cc}
  \toprule
  \bfseries Variables & \bfseries MCA\\
  \midrule
  true beliefs - hidden states & $\begin{bmatrix}
    0.958 & 0.355 & 0.466 & 0.071 & 0.990
  \end{bmatrix}$
  \\
  \addlinespace[1.ex]
  pred beliefs - hidden states & $\begin{bmatrix}
    0.801 & 0.255 & 0.611 & 0.000 & 1.000
  \end{bmatrix}$\\
  \bottomrule
  \end{tabular}}
\end{table}

It is also important to highlight that, while learning the optimal policy in this discrete benchmark might be considered straightforward, the POMDP/belief inference task is no small feat. It is perhaps more complex than in the continuous benchmark, given the non-identifiability of the learning problem. For instance, we cannot prevent the model from learning a different meaning of the hidden states (e.g., in the benchmark $s_0$ is the best hidden state, but this can be assigned to a different hidden state by the model). In general, many different solutions are possible and equivalent, which makes the learning process and the convergence more challenging. Nevertheless, the \emph{DBMM} is still able to learn a belief representation (and, hence, a POMDP model) that is nearly as good as the true one.

\subsection{Continuous control benchmark}
\label{sec:cont_benchmark}

The continuous control benchmark, where hidden states, actions, and observations are assumed continuous, corresponds to a setting for which the \emph{DBMM} is ideally suited, given that other methods, such as based on MCMC inference, become computationally infeasible. However, it is no longer possible to analytically compute the true beliefs and, thus, compare them against the beliefs estimated by the \emph{DBMM}. 

The benchmark problem examined herein is inspired by Sch{\"o}bi et al. \cite{schobi2016maintenance}, with some minor modifications. The primary difference lies in use of continuous observations, and an associated observation generating process, that replace the originally assumed discrete observations. Both the transition dynamics $\mathbb{P}$ and the observation generating process $\mathbb{O}$ are non-linear functions with Gaussian likelihoods. The details of the continuous control benchmark are reported in \ref{apd:cont}.

As earlier, all neural networks that compose the \emph{DBMM} comprise a single hidden layer of 100 neurons and a minimal hyper-parameter optimization was performed. Differently to the previous example, all neural networks learn now Gaussian distributions. In particular, the belief transition model is defined as:
\begin{equation}
    \tilde{b}_{\omega_t}=\mathcal{T}_\omega(\tilde{b}_{\omega_{t-1}}, a_{t-1})=\mathcal{N}(\mu_{\omega_{s_t}}, \sigma_{\omega_{s_t}})
\end{equation}
with $s_{\omega_t}\sim\tilde{b}_{\omega_t}$. The observation model:
\begin{equation}
    \mathbb{O}_\kappa(s_{\omega_t})=p_\kappa(o_t\mid s_{\omega_t})=\mathcal{N}(\mu_{\kappa_{o_t}}, \sigma_{\kappa_{o_t}})
\end{equation}
with $o_{\kappa_t}\sim p_\kappa(o_t\mid s_{\omega_t})$. Finally, the belief inference model:
\begin{equation}
    b_{\phi_t}=\mathcal{Q}_\psi(\tilde{b}_{\phi_{t}}, o_{t})=\mathcal{N}(\mu_{\psi{s_t}}, \sigma_{\psi{s_t}})
\end{equation}
Thus, the beliefs $\tilde{b}_{\omega_t}$ and $b_{\phi_t}$ are both represented by Gaussian distributions, with the structural form parameterized by non-linear neural networks. Note that, as explained in the previous section, the \emph{DBMM} is not limited to the Gaussian assumption but also non-Gaussian distributions and/or even mixtures can be modeled.

The evaluation procedure is the same as outlined in \algorithmref{alg:evaluation}, with the difference that true beliefs cannot be computed and, hence, these cannot be straightforwardly compared against the estimated beliefs. Since the true observation generating process is a Gaussian distribution centered in the true hidden state, and its variance is a non-linear function of the previous true hidden state (see \ref{apd:cont}), the observations are in effect noisy values of the hidden states. Therefore, to assess the accuracy of the \emph{DBMM} estimates, we compare the difference between the (environment) true observations and the true hidden states, with the difference between the estimated beliefs (we employ the estimated mean $\mu_{\psi{s_t}}$) and the true hidden states. 

As the observations are centered in the hidden states, learning beliefs with a smaller error than the observations is a challenging task. This stems from the unsupervised nature of the task as well as having only one realization per trial available, which renders learning the noise more difficult. If a lower error is achieved, this would allow the decision maker to base the decisions on a variable that is closer to the true hidden states than the available observations and, thus, to bring improved results. In addition, as no gold standard for the beliefs is available, and hence no known bound on the \emph{DBMM} performance exists, we further evaluate the (mean) belief predictions of an Ensemble Kalman Filter (EnKF) \cite{Evensen2009,katzfuss2016understanding}. This model assumes perfect knowledge of the problem, i.e., it uses the true POMDP model to compute the belief updates. Combining this with the problem setting, e.g., Gaussian likelihood of the transition dynamics and of the observation generating process, and a sufficiently high number of particles - we adopt 1000 particles - the EnKF is expected to provide estimates that can be very close to the gold standard. It is important to note that this does not necessarily constitute a model comparison between the EnKF and the \emph{DBMM}, as the former assumes ground truth knowledge of the entire POMDP model formulation, while this is instead learned by the \emph{DBMM}.

\begin{figure}
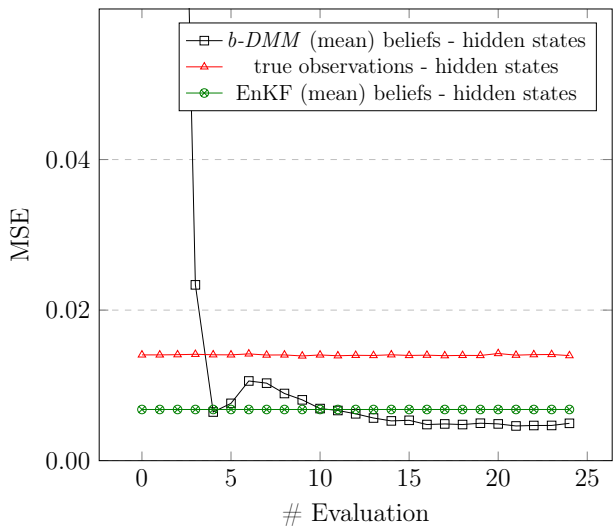

    \centering
    \includeteximage[width=0.6\textwidth]{Continuous_MSE_image_zoomed}
    \caption{MSE between the observations and the hidden states (red), MSE between the \emph{DBMM} (mean) beliefs and the hidden states (black), and MSE between the EnKF (mean) beliefs and the hidden states (green) on the continuous benchmark.}
    \label{fig:Continuous_MSE_image_zoomed}
\end{figure}

\figureref{fig:Continuous_MSE_image_zoomed} displays the Mean Squared Error (MSE) between the true hidden states and the observations, the MSE between the true hidden states and the estimated mean of the \emph{DBMM} beliefs, and the MSE between the true hidden states and the estimated mean of the EnKF beliefs. At the first evaluation, prior to any model training, the \emph{DBMM} predictions are random, but, after a few model updates, the estimated beliefs already outperform the observations, with their mean largely closer to the hidden states. As such, the \emph{DBMM} is able to learn the POMDP structure and a better representation of the hidden states than the available observations. Additionally, the \emph{DBMM} predictions eventually converge to a solution that is even better than the EnKF, albeit without assumption on the true POMDP model knowledge.

\begin{figure}
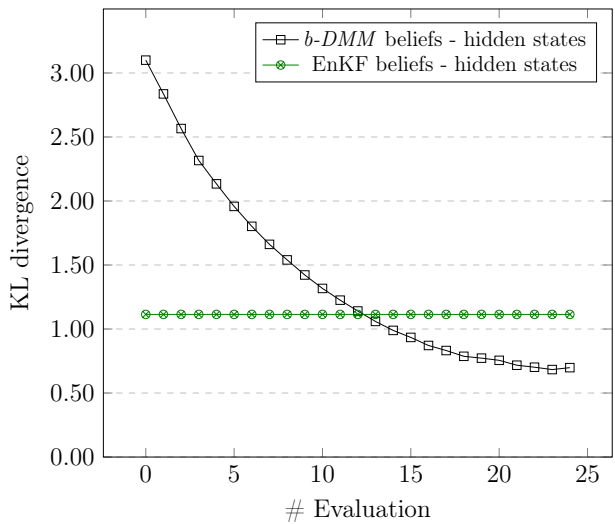

    \centering
    \includeteximage[width=0.6\textwidth]{Continuous_KL_image}
    \caption{KL divergence between \emph{DBMM} beliefs and the hidden states (black), and KL divergence between the EnKF beliefs and the hidden states (green).}
    \label{fig:Continuous_KL_image}
\end{figure}

While the former evaluation is important for confirming that the \emph{DBMM} is indeed able to provide predictions that are closer to the true hidden states than the available observations, this is somewhat limited as only the mean prediction is being evaluated. As such, we provide a further evaluation for the entire estimated belief distribution. Figure \ref{fig:Continuous_KL_image} displays the KL divergence between the \emph{DBMM} beliefs and the true hidden state distributions,
and the KL divergence between the EnKF
beliefs and the true hidden state distributions. As seen, the \emph{DBMM} is not only able to estimate belief distributions with a very accurate mean, but also the predicted variance $\sigma_{\psi{s_t}}$ - as also learned by neural networks in a completely unsupervised fashion - is very close to the true variance of the hidden state distributions. 

\begin{figure}
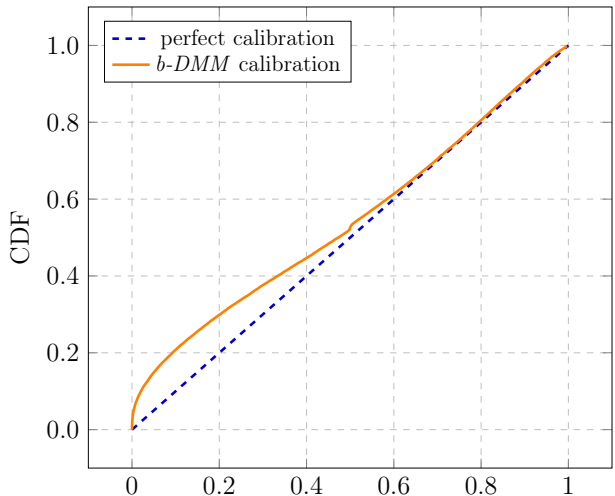

    \centering
    \includeteximage[width=0.6\textwidth]{reliability_diagram}
    \caption{Reliability diagram of \emph{DBMM} beliefs.}
    \label{fig:reliability_diagram}
\end{figure}

Lastly, we investigate whether the \emph{DBMM} belief estimates are well-calibrated. Calibration \cite{gneiting2007probabilistic} refers to the statistical consistency between the estimated distribution and the underlying true variable, e.g., the true labels should fall in a 90\% confidence interval approximately 90\% of the time. A diagnostic tool to assess the calibration performance is the reliability (or calibration) diagram \cite{gneiting2007probabilistic,kuleshov2018accurate}. In our problem setting with forecasts of continuous variables, the reliability diagram corresponds to the empirical Cumulative Distribution Function (CDF) of the estimated beliefs at the true hidden state values. Perfectly calibrated beliefs should be uniformly distributed, i.e., the empirical CDF should be close to the diagonal. Figure \ref{fig:reliability_diagram} displays the reliability diagram of the beliefs estimated by the \emph{DBMM}. The results, which demonstrate that the estimated beliefs lie indeed close to the diagonal, are particularly significant when considering the unsupervised nature of our task. Note that the calibration itself does not mean that the predictions are accurate, but along with the previous evaluations it is possible to conclude that the \emph{DBMM} belief forecasts are accurate and well-calibrated.

\subsection{Railway maintenance planning benchmark}

As a final evaluation benchmark, a railway maintenance planning problem is employed. The benchmark is characterized by discrete (railway track condition) hidden states and (maintenance) actions, $|\set{S}|=4$ and $|\set{A}|=3$, and continuous negative-valued observations, $|\set{O}|=\mathbb{R}^-$, which correspond to a track condition signal extracted from a laser-based monitoring system mounted on a diagnostic vehicle. The observation generation process is modeled via an autoregressive Truncated Student's $t$ processes:
\begin{align}\label{eq:obs-model}
\begin{split}
    z_t\mid t=0  & \sim \TS(\mu_{s_{t_0}}, \sigma_{s_{t_0}}, \nu_{s_{t_0}}, \ub=0)\\
    z_t - z_{t-1}\mid t > 0, a_{t-1}=0 & \sim \TS(\mu_{d\mid s_t}, \sigma_{d\mid s_t}, \nu_{d\mid s_t}, \ub=- z_{t-1})\\
    z_t\mid t > 0, a_{t-1}\in \{1,2\}  & \sim \TS(k_{r\mid a_{t-1}}*z_{t-1}+\mu_{r\mid s_t}, \sigma_{r\mid s_t}, \nu_{r\mid s_t}, \ub=0)\\
\end{split}
\end{align}
The transition and the observation models have been inferred from real-world data of monitoring measurements and maintenance actions collected over a decade on the Swiss railway network, allowing to reproduce a simulated environment that resembles the real-world problem with high-fidelity.
The interested reader is referred to Arcieri et al. \cite{arcieri2023bridging} for further details on the problem.

Despite the associated continuous observations and the challenging (to learn) generating process, the discrete states and actions allow to compute the true beliefs analytically via Equation \ref{eq:belief_discrete}. As such, the \emph{DBMM} belief estimates can directly be compared to the true beliefs here, without the need for further evaluations and gold standard approximations, as performed in the previous section. The \emph{DBMM} is composed of one-single hidden layer neural networks, which parameterize probability distributions as explained in the following. In particular, the belief transition model learns Categorical distributions, similarly to Section \ref{sec:disc_benchmark}: 
\begin{equation}
\tilde{b}_{\omega_t}=\mathcal{T}_\omega(\tilde{b}_{\omega_{t-1}}, a_{t-1})
\end{equation}
The belief inference model also learns Categorical distributions, with the difference that $o_{t-1}$ is additionally passed as input, given the autoregressive nature of the observations, namely:
\begin{equation}
    b_{\phi_t}=\mathcal{Q}_\psi(\tilde{b}_{\phi_{t}}, o_{t}, o_{t-1})
\end{equation}
Finally, the \emph{DBMM} observation model parameterize autoregressive Truncated Normal distributions, namely:
\begin{equation}
    \mathbb{O}_\kappa(s_{\omega_t}, a_{t-1}, o_{t-1})=p_\kappa(o_t\mid s_{\omega_t}, a_{t-1}, o_{t-1})=\TN(\mu_{\kappa_{o_t}}, \sigma_{\kappa_{o_t}}, \ub=0)
\end{equation}
with $o_{\kappa_t}\sim p_\kappa(o_t\mid s_{\omega_t}, a_{t-1}, o_{t-1})$. Note that the \emph{DBMM} observation model parameterizes Truncated Normal distributions, although the true generating process is modeled with a Truncated Student's \emph{t} distribution. The reason for this deviation lies in the lack of a straightforward implementation of the truncated Student's \emph{t} distribution in Pyro \cite{bingham2019pyro}, which is used for the \emph{DBMM} implementation. While this is a real problem and is expected to be more challenging, this discrepancy further increases the challenge for neural network learning.

The evaluation procedure is performed as in Algorithm \ref{alg:evaluation} in \ref{apd:alg} with $T=50$ time-steps, as in the original problem. Table \ref{tab:mca-rail} reports the MCA and the KL divergence of i) true beliefs with respect to the true hidden states and of ii) predicted beliefs with respect to the true hidden states. Despite the increased complexity of this problem, the \emph{DBMM} is still able here to learn a valuable belief representation, providing meaningful information over all classes of hidden states, when compared against the ground truth beliefs.

\begin{table}[hbtp]
\floatconts
  {tab:mca-rail}
  {\caption{Multi-Class Accuracy (MCA), i.e., per-class accuracy score, and KL divergence of the true beliefs with respect to the true hidden states (top row) and of the predicted beliefs with respect to the true hidden states (bottom row).}}
  {\begin{tabular}{ccc}
  \toprule
  \bfseries Variables & \bfseries MCA  & \bfseries KL loss\\
  \midrule
  true beliefs - hidden states & $\begin{bmatrix}
    0.934 & 0.875 & 0.760 & 0.893
  \end{bmatrix}$ & 0.89
  \\
  \addlinespace[1.ex]
  pred beliefs - hidden states & $\begin{bmatrix}
    0.800 & 0.477 & 0.608 & 0.665
  \end{bmatrix}$ & 1.28 \\
  \bottomrule
  \end{tabular}}
\end{table}

\section{Conclusion}
\label{sec:conclusion}
This work introduces the Deep Belief Markov Model (\emph{DBMM}), an extension of the DMM specifically designed for application in POMDP and belief inference. The \emph{DBMM} allows a model-formulation agnostic POMDP and belief structured inference based on variational learning. By leveraging neural networks, the \emph{DBMM} is able to handle complex input structures, e.g., continuous and/or multi-dimensional POMDP variables with possibly non-linear system dynamics, which would complicate and/or preclude application of existing methods, such as those based, for example, on MCMC sampling. 

In this work, the \emph{DBMM} inference is benchmarked on three POMDP problems: one  with discrete variables, for which the true beliefs can be analytically computed and compared against the model-estimated beliefs; one with continuous variables, for which true beliefs can no longer be analytically computed and suited inference methods are lacking; and one comprising a mixture of discrete and continuous variables, represented by a real-world problem of railway optimal maintenance, with a particularly challenging observation generating process. The proposed model is able to learn the underlying POMDP model and a useful belief representation in all benchmarks. Specifically, the \emph{DBMM} is shown able to learn network belief representations of the hidden states that i) eventually converge to the ground truth belief representations in the discrete state POMDP cases, and ii) provide better information than the available observations and than an EnKF implementation in the continuous POMDP case, despite the EnKF assuming ground truth knowledge of the POMDP problem for the belief updates.

Overall, the \emph{DBMM} offers a highly promising direction for inference and toward improving decision-making in POMDP problems. By exclusively relying on available observations and actions taken, the \emph{DBMM} is able to estimate accurate and well-calibrated beliefs. Follow-up work will combine the POMDP/RL training with the \emph{DBMM} to demonstrate the application of this method and reveal its merits in learning a policy of a system, which may also be changing in time, either due to performed actions or a natural/physical process, and be continuously inferred by the \emph{DBMM}.

A possible limitation of the \emph{DBMM} presented in this work is that, while essentially no assumptions are made on the structural form of the transition dynamics and the observation generating process, the likelihood form was assumed to be known in the continuous POMDP and in the railway benchmark (Gaussian and Truncated Normal, respectively). Future work can combine the proposed model with autoregressive flows \cite{rezende2015variational,kingma2016improved} to relax this assumption.

\appendix
\section{Problem description}\label{apd:problems}

\subsection{Discrete control problem}\label{apd:disc}

The discrete control problem resembles a bridge maintenance planning problem, modeled as a POMDP, applied in Papakonstantinou et al. \cite{papakonstantinou2018pomdp} and originally presented in Corotis et al. \cite{corotis2005modeling}. The problem presents variable spaces with dimensionality $|\set{S}|=5$, $|\set{A}|=4$, and $|\set{O}|=3$. The transition matrices for each of the 4 actions, as well as the observation likelihoods are given in the following:

\[
\mathbb{P}_1=
  \begin{bmatrix}
    0.80 & 0.13 & 0.02 & 0.00 & 0.05 \\
    0.00 & 0.70 & 0.17 & 0.05 & 0.08 \\
    0.00 & 0.00 & 0.75 & 0.15 & 0.10 \\
    0.00 & 0.00 & 0.00 & 0.60 & 0.40 \\
    0.00 & 0.00 & 0.00 & 0.00 & 1.00
  \end{bmatrix}
\]

\[
\mathbb{P}_2=
  \begin{bmatrix}
    0.80 & 0.13 & 0.02 & 0.00 & 0.05 \\
    0.00 & 0.80 & 0.10 & 0.02 & 0.08 \\
    0.00 & 0.00 & 0.80 & 0.10 & 0.10 \\
    0.00 & 0.00 & 0.00 & 0.60 & 0.40 \\
    0.00 & 0.00 & 0.00 & 0.00 & 1.00
  \end{bmatrix}
\]

\[
\mathbb{P}_3=
  \begin{bmatrix}
    0.80 & 0.13 & 0.02 & 0.00 & 0.05 \\
    0.19 & 0.65 & 0.08 & 0.02 & 0.06 \\
    0.10 & 0.20 & 0.56 & 0.08 & 0.06 \\
    0.00 & 0.10 & 0.25 & 0.55 & 0.10 \\
    0.00 & 0.00 & 0.00 & 0.00 & 1.00
  \end{bmatrix}
\]

\[
\mathbb{P}_4=
  \begin{bmatrix}
    0.80 & 0.13 & 0.02 & 0.00 & 0.05 \\
    0.80 & 0.13 & 0.02 & 0.00 & 0.05 \\
    0.80 & 0.13 & 0.02 & 0.00 & 0.05 \\
    0.80 & 0.13 & 0.02 & 0.00 & 0.05 \\
    0.80 & 0.13 & 0.02 & 0.00 & 0.05
  \end{bmatrix}
\] 

\[
\mathbb{O}=
  \begin{bmatrix}
    0.80 & 0.20 & 0.00 \\
    0.20 & 0.60 & 0.20 \\
    0.05 & 0.70 & 0.25 \\
    0.00 & 0.30 & 0.70 \\
    0.00 & 0.00 & 1.00
  \end{bmatrix}
\]

\subsection{Continuous control problem}\label{apd:cont}

The continuous control problem is inspired by the continuous (hidden states) maintenance planning problem in Sch{\"o}bi et al. \cite{schobi2016maintenance}, where the transition dynamics and observation generating process have been changed to further include continuous actions and observations (instead of discrete variables). The variables spaces are thus continuous, specifically $\set{S}=\mathbb{R}$, $\set{A}=[0, 1]$, and $\set{O}=\mathbb{R}$. 

The transition dynamics has the structural form:

\begin{equation}
    s_{t+1} = f_{a_t}(s_t) + \sigma_{a_t}(s_t)\cdot\mathcal{N}(0,1) 
\end{equation}
namely, the mean and standard deviation of the next state depend on non-linear functions $f_{a_t}$ and $\sigma_{a_t}$, respectively, of the current state, which in turn depend on the action $a_t$. For action $a_t=0$ (do-nothing, pure deterioration process):
\begin{equation}
  \begin{aligned}
    f_{det}(s_t) & = \max(0, s_t - \exp(-5s_t)\cdot0.5 -1)\\
    \sigma_{det}(s_t) & = \frac{\max(0, s_t)-\max(0, f_{det}(s_t))}{2}+0.02
  \end{aligned}
\end{equation}
For action $a_t=1$ (full replace):
\begin{equation}
  \begin{aligned}
    f_{rep}(s_t) & = 0.96\\
    \sigma_{det}(s_t) & = 0.02
  \end{aligned}
\end{equation}
The overall dynamics for actions $a_t\in\set{A}$ of intermediate intensity is given by the combination of the two processes:
\begin{equation}
    \mathbb{P}=f_{rep}(s_t)\cdot a_t + f_{det}(s_t)\cdot(1-a_t)
\end{equation}

\begin{figure}[htbp]
    \centering
    \begin{tikzpicture}
    \node (img)  {\includegraphics[width=0.55\textwidth]{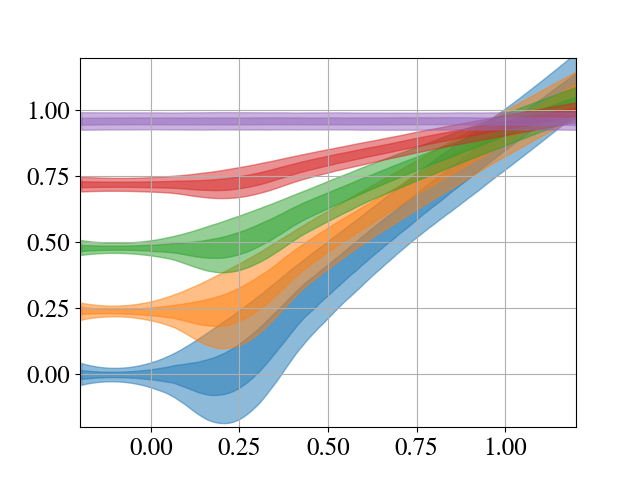}};
    \node[below=of img, node distance=0cm,xshift=0cm,  yshift=1.2cm] {$s_t$};
    \node[left=of img, node distance=0cm, rotate=90, anchor=center,xshift=0cm,yshift=-1cm] {$s_{t+1}$};
    \end{tikzpicture}
    \caption{Transition dynamics of the continuous environment for 5 different actions, namely $a_t\in\{0, 0.25, 0.5, 0.75, 1\}$, reported in blue, orange, green, red, and purple, respectively. The $50\%$ (darker) and $90\%$ (lighter) distribution of the next state is shown for all actions.}
    \label{fig:cont_env}
\end{figure}

An explanatory representation of the dynamics is reported in Figure \ref{fig:cont_env} for 5 different actions, namely $a_t\in\{0, 0.25, 0.5, 0.75, 1\}$ in blue, orange, green, red, and purple, respectively. Every curve shows the $50\%$ (darker) and $90\%$ (lighter) distribution of the next state given $s_t$.

The continuous observations are generated by a Gaussian likelihood centered in the states, with variance that is a non-linear function of the state, namely:
\begin{equation}
    o_t \sim \mathcal{N}(s_t, 0.005\exp(s_t))
\end{equation}
In particular, the noise of the observations is highest at a perfect condition, i.e., $s_t\geq 1$, and smallest at the worst condition, i.e., $s_t\leq 0$, reflecting the realistic feature that the failure condition is detected with more certainty.

\newpage

\section{Evaluation algorithm}\label{apd:alg}

The evaluation procedure is reported in Algorithm \ref{alg:evaluation}:

\begin{algorithm}
\caption{Evaluation Deep Belief Markov Model}
\label{alg:evaluation}
\KwIn{$\pi, \tilde{b}_0$}
\KwOut{Evaluation of \emph{DBMM}}
\For{evaluation $i\leftarrow 1$ \KwTo $N$}{
   \For{trial $k\leftarrow 1$ \KwTo $500$}{
      $o_0\leftarrow\texttt{env.reset()}$\;
      $b_{\phi_0}\leftarrow\texttt{b\_dmm.inference(}\tilde{b}_0, o_0\texttt{)}$ \tcp*{Equation \ref{eq:inf_bdmm}}\
      $a_0\sim \pi$\;
      \For{time-step $t\leftarrow 1$ \KwTo $100$}{
         $o_t\leftarrow\texttt{env.step(}a_{t-1}\texttt{)}$\;
         $b_{\phi_t}\leftarrow\texttt{b\_dmm.inference(}b_{\phi_{t-1}}, o_t, a_{t-1}\texttt{)}$ \tcp*{Equation \ref{eq:t_bdmm} and \ref{eq:inf_bdmm}}\
         $s_t, b_t\leftarrow\texttt{env.true\_state\_and\_belief()}$\tcp*{Belief computed via Equation \ref{eq:belief_discrete}}\
         $a_{t}\sim \pi$\;
      }
   }
  Run evaluation of $b_{\phi_{1:100}}, b_{1:100}, s_{1:100}\;\forall\;trials\;k$\;
  Update \texttt{b\_dmm} with $o_{1:100}, a_{0:99}\;\forall\;trials\;k$ \tcp*{Equation \ref{eq:vlb_bdmm}}\
}
\end{algorithm}


\newpage
\bibliographystyle{elsarticle-num} 
\bibliography{main.bib}

\end{document}